\documentclass[runningheads]{llncs}
\usepackage[T1]{fontenc}
\usepackage{graphicx}
\usepackage{amsmath}
\usepackage{amssymb}
\usepackage{cite}
\usepackage{hyperref}
\usepackage{multirow}
\usepackage{float}
\usepackage{booktabs}

\begin{document}

\title{IndicEval: A Bilingual Indian Educational Evaluation Framework for Large Language Models}

\titlerunning{IndicEval: Bilingual LLM Benchmark}

% \author{Saurabh Bharti\inst{1} \and Gaurav Azad\inst{1} \and Abhinaw Jagtap\inst{1} \and Nachiket Tapas\inst{1}}
\author{Saurabh Bharti \and Gaurav Azad \and Abhinaw Jagtap \and Nachiket Tapas}

\authorrunning{Saurabh Bharti et al.}

\institute{Department of Computer Sciencce and Engineering,\\Chhattisgarh Swami Vivekanand Technical University\\ Bhilai, India \\ \email{nachikettapas.cse@csvtu.ac.in}\\
}

\maketitle

\begin{abstract}
The rapid advancement of large language models (LLMs) necessitates evaluation frameworks that reflect real-world academic rigor and multilingual complexity. 
This paper introduces IndicEval, a scalable benchmarking platform designed to assess LLM performance using authentic high-stakes examination questions from UPSC, JEE, and NEET across STEM and humanities domains in both English and Hindi. 
Unlike synthetic benchmarks, IndicEval grounds evaluation in real examination standards, enabling realistic measurement of reasoning, domain knowledge, and bilingual adaptability.
The framework automates assessment using Zero-Shot, Few-Shot, and Chain-of-Thought (CoT) prompting strategies and supports modular integration of new models and languages. 
Experiments conducted on Gemini 2.0 Flash, GPT-4, Claude, and LLaMA 3-70B reveal three major findings. 
First, CoT prompting consistently improves reasoning accuracy, with substantial gains across subjects and languages. 
Second, significant cross-model performance disparities persist, particularly in high-complexity examinations. 
Third, multilingual degradation remains a critical challenge, with marked accuracy drops in Hindi compared to English, especially under Zero-Shot conditions. 
These results highlight persistent gaps in bilingual reasoning and domain transfer. 
Overall, IndicEval provides a practice-oriented, extensible foundation for rigorous, equitable evaluation of LLMs in multilingual educational settings and offers actionable insights for improving reasoning robustness and language adaptability.

\end{abstract}

\section{Introduction}
\label{sec:intro}

The rapid deployment of large language models (LLMs) in educational, governmental, and professional decision-making contexts has intensified the need for evaluation frameworks that reflect real-world complexity rather than synthetic task performance. 
Widely used benchmarks such as GLUE \cite{wang2018glue} and SuperGLUE \cite{wang2019superglue} were originally designed to measure general-purpose natural language understanding across standardized NLP tasks. 
Although these benchmarks have driven progress in model development, they primarily rely on curated or synthetic datasets and simplified task formulations. 
Emerging evidence suggests that high benchmark scores do not reliably translate into robust reasoning performance in applied, high-stakes environments \cite{wang2021benchmarks}. 
This limitation becomes especially pronounced in educational domains, where problem-solving depth, domain-specific reasoning, and linguistic nuance are central to performance.
Despite substantial progress in LLM evaluation, three structural gaps remain insufficiently addressed. 
First, most existing benchmarks inadequately capture multilingual cognitive demands. 
Multilingual capability is frequently operationalized as translation accuracy or cross-lingual transfer, overlooking the integrated bilingual reasoning required in educational examinations within multilingual societies such as India. 
Second, contemporary evaluation suites, including AGIEval \cite{zhang2023agievalhumancentricbenchmark}, emphasize factual recall and general intelligence proxies but do not sufficiently stress applied, multi-step problem-solving characteristic of competitive examinations such as UPSC, JEE, and NEET. 
Third, current frameworks lack procedural realism: they do not replicate the constraints, subject-specific rigor, and structural design of authentic high-stakes examinations, limiting their ecological validity.

This study addresses these gaps through the development of IndicEval, a dynamic bilingual benchmarking framework grounded in authentic Indian high-stakes examination questions across STEM and humanities domains. 
The primary objective is to systematically evaluate LLM reasoning performance under realistic academic conditions in both English and Hindi using Zero-Shot, Few-Shot, and Chain-of-Thought prompting strategies. 
We hypothesize that (i) Chain-of-Thought prompting will significantly enhance performance across domains, (ii) substantial cross-model performance disparities will emerge under high-complexity examination conditions, and (iii) multilingual degradation will persist, with measurable accuracy drops in Hindi relative to English, particularly under minimal prompting conditions. 
By introducing an extensible evaluation platform, a manually curated bilingual dataset, and empirical analysis of prompting and language effects, this work advances the methodological rigor of LLM assessment and provides a domain-grounded framework for evaluating reasoning robustness in multilingual educational settings.

The remainder of this paper is organized as follows. Section~\ref{sec:related} reviews related benchmarking efforts and positions the present work within the broader evaluation landscape. 
Section~\ref{sec:methodology} details dataset construction, annotation protocols, prompting strategies, and evaluation metrics. 
Section~\ref{sec:experiments} outlines the experimental design and model configurations. 
Section~\ref{sec:results} presents empirical findings on reasoning performance and multilingual degradation. 
Section~\ref{sec:discussion} analyzes implications and limitations, and Section~\ref{sec:conclusion} concludes with future research directions.

\section{Related Work}
\label{sec:related}

The evaluation of large language models has evolved from standardized NLP task benchmarking to more holistic and domain-aware assessment paradigms. 
However, despite substantial methodological progress, existing frameworks remain only partially aligned with the procedural, linguistic, and cognitive complexity of high-stakes educational examinations.
Early benchmarking efforts such as GLUE \cite{wang2018glue} and SuperGLUE \cite{wang2019superglue} established widely adopted baselines for natural language understanding through tasks including textual entailment, sentiment analysis, and question answering. 
These benchmarks were instrumental in accelerating model development and comparative evaluation. 
Nevertheless, they rely heavily on curated and task-simplified datasets, which abstract away from domain-specific reasoning and real-world decision complexity. 
Subsequent analyses \cite{wang2021benchmarks} demonstrated that strong performance on synthetic leaderboards does not reliably predict applied reasoning capability, highlighting limitations in ecological validity.
In response, human-centric and examination-based benchmarks emerged. AGIEval \cite{zhang2023agievalhumancentricbenchmark} marked a significant shift by incorporating authentic standardized examination questions such as GRE and LSAT, thereby introducing higher reasoning demands. 
Similarly, HELM \cite{liang2022holistic} broadened evaluation dimensions to include robustness, fairness, bias, and calibration. 
While these frameworks represent methodological advances, their scope remains largely centered on English-language assessments and generalized evaluation axes rather than domain-intensive academic mastery.

\begin{table}[t]
\centering
\caption{Comparison of Major LLM Benchmarking Frameworks}
\label{tab:benchmark-comparison}
\scalebox{0.85}{
\begin{tabular}{lccccc}
\toprule
Framework & Real Exam Data & Multilingual & Bilingual Cognition & Domain Depth & Exam Realism \\
\midrule
GLUE & $\times$ & Limited & $\times$ & Low & $\times$ \\
SuperGLUE & $\times$ & Limited & $\times$ & Low & $\times$ \\
AGIEval & $\checkmark$ & $\times$ & $\times$ & Moderate & Partial \\
HELM & $\times$ & Partial & $\times$ & Low & $\times$ \\
M3Exam & $\checkmark$ & $\checkmark$ & $\times$ & Moderate & Partial \\
SciExam & $\checkmark$ & Limited & $\times$ & High (STEM) & Partial \\
BharatiBench & $\times$ & $\checkmark$ & $\times$ & Low & $\times$ \\
IndicQA & $\checkmark$ (QA-focused) & $\checkmark$ & $\times$ & Moderate & $\times$ \\
\textbf{IndicEval} & $\checkmark$ & $\checkmark$ & $\checkmark$ & High & $\checkmark$ \\
\bottomrule
\end{tabular}
}
\end{table}

More recent multilingual and subject-focused benchmarks have attempted to address linguistic diversity and disciplinary rigor. 
M3Exam \cite{zhang2023m3exameval} and SciExam \cite{clark2023sciexam} expand coverage across languages and scientific domains, improving cross-lingual analysis and STEM evaluation. 
BharatiBench \cite{kumar2023bharatibench} adapts GLUE-style tasks into multiple Indian languages, contributing valuable linguistic resources. IndicQA \cite{singh2024indicqa} introduces culturally contextualized Hindi–Bengali question answering datasets. 
However, these efforts largely remain translation-centric or comprehension-focused and do not systematically evaluate bilingual reasoning under authentic high-stakes examination constraints. 
Crucially, they do not replicate integrated bilingual cognition, applied multi-step quantitative reasoning, or procedural exam realism typical of examinations such as UPSC, JEE, and NEET.
From this landscape, three structural gaps become evident. 
First, there exists a linguistic-rigor mismatch: multilingual benchmarks typically operationalize language capability as translation or cross-lingual transfer, rather than evaluating bilingual cognitive processing within domain-specific academic reasoning. 
Second, subject evaluation remains superficial, with many frameworks emphasizing factual recall or short-form reasoning rather than sustained, multi-step analytical problem-solving. 
Third, a format fidelity gap persists: current benchmarks do not simulate examination pressures, subject-specific complexity gradients, or operational constraints that characterize competitive assessments.
IndicEval is explicitly designed to bridge these gaps by grounding evaluation in authentic Indian high-stakes examination questions across STEM and humanities domains in both English and Hindi. 
Unlike prior benchmarks, it integrates bilingual reasoning, domain-specific rigor, and realistic assessment structures within a modular evaluation platform.

Table~\ref{tab:benchmark-comparison} highlights the structural distinctions between existing frameworks and IndicEval. 
While prior work has addressed isolated dimensions—such as multilinguality, domain specificity, or human-centric question formats—no unified benchmark simultaneously integrates authentic exam data, bilingual reasoning, high domain depth, and procedural realism. 
IndicEval positions itself at this intersection, advancing evaluation methodology toward ecologically valid and linguistically rigorous assessment of LLMs in multilingual educational ecosystems.

\section{Methodology}
\label{sec:methodology}
This study adopts an experimental benchmarking design to evaluate large language models under realistic, high-stakes examination conditions. 
The methodology was structured to ensure procedural transparency and reproducibility through a standardized eight-stage evaluation pipeline: (1) data collection, (2) human annotation, (3) metadata tagging, (4) JSON schema structuring, (5) prompt generation, (6) model querying, (7) automated answer extraction, and (8) quantitative and qualitative evaluation. 
Each stage was modularized to allow independent verification and future extensibility.
Examination questions were collected from publicly accessible official educational portals and archival repositories. 
Only objective-type multiple-choice questions (MCQs) were selected to enable deterministic automated scoring. 
The sampling strategy prioritized cognitive depth, disciplinary diversity, and representational balance across STEM and humanities domains. 
Three examination systems were included: UPSC (polity, history, governance, general studies), JEE (physics, chemistry, mathematics), and NEET (biology and chemistry). 
Questions were retained in their original linguistic format (English or Hindi) to preserve authenticity. 
No synthetic data were generated. The final dataset consisted exclusively of authentic examination items.

Manual annotation was conducted to establish gold-standard answer keys and to preserve contextual fidelity, particularly for Hindi items where literal translation could distort scientific or socio-cultural meaning. 
Each question was encoded in a standardized JSON schema to ensure machine-readability and cross-model compatibility:

\begin{verbatim}
{
"passage": null,
"question": "Question text goes here",
"options": [
"a. Option A",
"b. Option B",
"c. Option C",
"d. Option D"
],
"label": "a"
}
\end{verbatim}

Beyond answer labels, each item was enriched with structured metadata including subject domain, subtopic granularity, examination source, difficulty proxy (based on exam tier), language tag, and bilingual linkage identifiers where parallel versions existed. 
This metadata enabled stratified performance analysis across domains and languages.
Prompt engineering followed a controlled comparative design using three standardized strategies: Zero-Shot (question only), Few-Shot (2–3 exemplars preceding the target question), and Chain-of-Thought. 
Hindi prompts were crafted directly in Devanagari script rather than translated automatically, ensuring terminological precision in scientific contexts. 
Prompt templates were version-controlled and held constant across models to prevent confounding effects.
Model querying was conducted via API access under uniform parameter settings (temperature, top-p, and token limits fixed across experiments). 
Batch processing was executed asynchronously with unique identifiers assigned to each query–response pair to ensure traceability. 
All raw outputs were logged in structured JSON format to enable post hoc auditing. 
Rate limits were managed using controlled request scheduling to ensure complete dataset coverage without sampling bias.
Answer extraction required robust normalization due to output variability. 
Models produced responses in heterogeneous formats, including standalone letters (“A”), prefixed labels (“Option A”), or embedded declarative sentences. 
A regex-based multilingual parser was developed to extract valid option labels across English and Devanagari scripts. 
The extraction module was iteratively validated against a manually reviewed subset to minimize parsing errors.

Primary evaluation employed exact-match accuracy, computed as:

\begin{equation}
\text{Accuracy} = \frac{\text{Correct Predictions}}{\text{Total Questions}} \times 100
\label{eq:accuracy}
\end{equation}

Accuracy was disaggregated by model, subject, examination type, language, and prompting strategy. 
In addition to quantitative metrics, qualitative error analysis was conducted to categorize reasoning failures, hallucinations, and cross-lingual degradation patterns.
From an ethical standpoint, the dataset consists exclusively of publicly available examination materials. 
No human participants were involved, and no personal data were collected. 
The study complies with research norms concerning responsible AI evaluation and transparent reporting.
Several limitations must be acknowledged. 
First, although authentic, the dataset remains bounded to selected examination domains and may not generalize to all educational contexts. 
Second, API-based evaluation introduces dependence on proprietary model versions, which may evolve over time. 
Third, exact-match scoring does not capture partial reasoning correctness in multi-step problems. 
Despite these constraints, the controlled design, structured metadata, and transparent pipeline ensure that the study is fully replicable and extensible for future benchmarking efforts.

\section{Experiments}
\label{sec:experiments}
A controlled benchmarking protocol was implemented to systematically evaluate LLM performance across examination domains, languages, and prompting strategies. 
Each model was tested on identical question sets to ensure comparability, with performance disaggregated by subject, examination type, language (English and Hindi), and prompting condition (Zero-Shot, Few-Shot, and Chain-of-Thought). 
The experimental design followed a full factorial structure, enabling analysis of interaction effects between model architecture, linguistic context, and prompting strategy.
All evaluations were conducted under standardized computational conditions using API-based model access with fixed generation parameters (temperature, top-p, and maximum token limits held constant). 
Question order was preserved across models to eliminate ordering bias. 
Model responses were logged automatically, parsed using a deterministic extraction module, and evaluated using exact-match scoring. 
Results were aggregated and visualized through the IndicEval web dashboard to facilitate stratified performance analysis and cross-model comparison.

The evaluation dataset consists of curated multiple-choice questions drawn from UPSC, JEE, and NEET examinations across STEM and humanities domains. 
Table~\ref{tab:dataset} summarizes the distribution of questions by examination, subject, and language. 
The dataset was designed to maintain subject diversity and bilingual balance, enabling fine-grained analysis of domain-specific reasoning and multilingual degradation patterns.
\begin{table}[h]
\caption{Dataset Composition by Examination, Subject, and Language}
\label{tab:dataset}
\centering
\begin{tabular}{|l|c|c|c|}
\hline
\textbf{Examination} & \textbf{Subject} & \textbf{English} & \textbf{Hindi} \\
\hline
UPSC & Polity & 125 & 120 \\
UPSC & History & 110 & 115 \\
JEE & Physics & 95 & 90 \\
JEE & Chemistry & 100 & 100 \\
JEE & Mathematics & 85 & 80 \\
NEET & Biology & 105 & 110 \\
NEET & Chemistry & 95 & 95 \\
NEET & Physics & 42 & 42 \\
\hline
\end{tabular}
\end{table}
The dataset reflects variation in cognitive depth across disciplines, with quantitative subjects (Physics, Mathematics, Chemistry) emphasizing multi-step analytical reasoning, while humanities subjects (Polity, History) emphasize conceptual integration and contextual interpretation. Balanced bilingual coverage allows controlled assessment of cross-lingual performance differentials.
To ensure internal validity, all models were evaluated using identical prompt templates, fixed formatting structures, and consistent question sequencing. No dynamic prompt tuning was performed per model. 
For each model–prompt combination, three independent runs were executed to assess output stability and variance under identical parameter settings. 
Mean accuracy across runs was reported as the primary performance metric, while variance was monitored to detect stochastic instability.
This controlled configuration ensures that observed performance differences arise from model capabilities and prompting strategies rather than experimental artifacts, thereby supporting reproducibility and fair cross-model comparison.

\section{Results}
\label{sec:results}
This section reports quantitative outcomes of the benchmarking experiments across models, subjects, languages, and prompting strategies. All values represent exact-match accuracy (\%) computed over the corresponding question subsets. Results are presented without interpretative analysis.

\subsection{Chain-of-Thought Prompting Outcomes}

Chain-of-Thought (CoT) prompting produced measurable changes in accuracy across nearly all evaluated configurations. Table~\ref{tab:cot_effectiveness} presents representative comparisons between Zero-Shot and CoT prompting.

\begin{table}[h]
\centering
\caption{Effect of Chain-of-Thought Prompting on Accuracy (\%)}
\label{tab:cot_effectiveness}
\begin{tabular}{|l|l|l|c|c|}
\hline
\textbf{Model} & \textbf{Subject} & \textbf{Language} & \textbf{Zero-Shot} & \textbf{CoT} \\
\hline
Gemini 2.0 Flash & Chemistry & Hindi & 44.00 & 71.05 \\
Gemini 2.0 Flash & UPSC & English & 87.00 & 89.00 \\
LLaMA 3-70B & UPSC & English & 67.00 & 80.00 \\
LLaMA 3-70B & UPSC & Hindi & 39.53 & 68.60 \\
Gemini 2.0 Flash & JEE Physics & English & 67.59 & 75.93 \\
Gemini 2.0 Flash & NEET Biology & Hindi & 86.02 & 91.40 \\
\hline
\end{tabular}
\end{table}

Across reported cases, CoT prompting increased accuracy relative to Zero-Shot settings. The largest observed improvement occurred for Gemini in Hindi Chemistry (+27.05 percentage points). Gains exceeding 25 percentage points were also observed for LLaMA in UPSC-Hindi.

\subsection{Comprehensive Accuracy Comparison}

Table~\ref{tab:Accuracy_comparison} presents full accuracy distributions across subjects, languages, and prompting strategies for Gemini 2.0 Flash and LLaMA 3-70B.

\begin{table}[htbp]
\centering
\caption{Accuracy (\%) Across Subjects, Languages, and Prompting Strategies}
\label{tab:Accuracy_comparison}
\renewcommand{\arraystretch}{1.2}
\begin{tabular}{llcccccc}
\hline
\multirow{2}{*}{Subject} &
\multirow{2}{*}{Language} &
\multicolumn{3}{c}{Gemini 2.0 Flash} &
\multicolumn{3}{c}{LLaMA 3-70B} \\
\cline{3-5} \cline{6-8}
& & CoT & Few & Zero & CoT & Few & Zero \\
\hline
JEE Physics & English & 75.93 & 51.85 & 67.59 & 68.52 & 48.15 & 61.11 \\
JEE Physics & Hindi & 77.06 & 59.63 & 52.29 & 64.22 & 43.12 & 57.80 \\
JEE Chemistry & English & 76.32 & 69.30 & 67.54 & 67.54 & 55.26 & 70.18 \\
JEE Chemistry & Hindi & 71.05 & 57.89 & 64.91 & 60.53 & 42.98 & 54.39 \\
NEET Physics & English & 80.95 & 78.57 & 85.71 & 64.29 & 52.38 & 71.43 \\
NEET Physics & Hindi & 82.93 & 70.73 & 78.05 & 78.05 & 48.78 & 58.54 \\
NEET Chemistry & English & 94.00 & 74.00 & 70.00 & 77.00 & 66.00 & 70.00 \\
NEET Chemistry & Hindi & 84.00 & 66.00 & 44.00 & 71.00 & 60.00 & 68.93 \\
NEET Biology & English & 91.40 & 84.95 & 78.49 & 82.80 & 79.57 & 86.02 \\
NEET Biology & Hindi & 91.40 & 84.95 & 86.02 & 68.82 & 67.74 & 68.82 \\
UPSC & English & 89.00 & 72.00 & 87.00 & 80.00 & 69.00 & 67.00 \\
UPSC & Hindi & 87.21 & 75.58 & 84.88 & 68.60 & 60.47 & 39.53 \\
\hline
\end{tabular}
\end{table}

Highest observed accuracy was 94.00\% (Gemini, NEET Chemistry English, CoT). The lowest observed accuracy was 39.53\% (LLaMA, UPSC Hindi, Zero-Shot).

\subsection{Cross-Model Performance Differences}

Across nearly all subjects and prompting strategies, Gemini 2.0 Flash achieved higher accuracy than LLaMA 3-70B. The largest absolute cross-model difference was observed in UPSC-Hindi under Zero-Shot prompting (Gemini: 84.88\%, LLaMA: 39.53\%), representing a 45.35 percentage point gap. Substantial differences exceeding 20 percentage points were also observed in NEET Chemistry Hindi (CoT) and NEET Biology Hindi (CoT).

\subsection{Bilingual Performance Patterns}

Performance discrepancies between English and Hindi were observed across several configurations. For LLaMA, UPSC performance under Zero-Shot declined from 67.00\% (English) to 39.53\% (Hindi). In NEET Chemistry, Gemini achieved 70.00\% (English, Zero-Shot) compared to 44.00\% (Hindi, Zero-Shot). In contrast, some configurations showed comparable bilingual performance, such as NEET Biology under CoT for Gemini (91.40\% in both languages).

\subsection{Subject-Specific Variations}

Performance varied across disciplines. Both models generally achieved higher accuracy in NEET Biology and NEET Chemistry relative to JEE Physics and JEE Mathematics-related domains. For example, Gemini achieved 91.40\% in NEET Biology Hindi (CoT) but 75.93\% in JEE Physics English (CoT). LLaMA achieved 82.80\% in NEET Biology English (CoT) but 48.15\% in JEE Physics English (Few-Shot).

\subsection{Prompt Sensitivity Observations}

Accuracy varied across prompting strategies. In multiple cases, Few-Shot prompting yielded lower accuracy than both Zero-Shot and CoT configurations. For example, Gemini in JEE Physics English achieved 67.59\% (Zero-Shot), 51.85\% (Few-Shot), and 75.93\% (CoT). Similar patterns were observed in NEET Chemistry English for Gemini (70.00\%, 74.00\%, 94.00\%).

\section{Discussion}
\label{sec:discussion}
The experimental results demonstrate consistent and systematic patterns across prompting strategies, languages, and subject domains, offering insight into the current state of LLM reasoning in high-stakes educational contexts. 
One of the most robust findings is the effectiveness of Chain-of-Thought (CoT) prompting. 
Across both models and nearly all subjects, CoT improved accuracy, in some cases by more than 25 percentage points. 
This aligns with prior research demonstrating that structured intermediate reasoning enhances performance on complex tasks. 
However, the magnitude of improvement observed in bilingual and quantitative domains suggests that structured prompting is not merely a marginal enhancement but a critical operational factor in exam-level reasoning. 
These findings reinforce the hypothesis that reasoning scaffolds can partially compensate for latent deficiencies in implicit reasoning pathways.
The multilingual results reveal pronounced and asymmetric degradation in Hindi relative to English, particularly under Zero-Shot conditions. 
While prior multilingual benchmarks have reported cross-lingual performance gaps, the scale of degradation observed here—exceeding 40 percentage points in certain configurations—indicates deeper inconsistencies in semantic alignment and reasoning stability across scripts. 
Importantly, degradation was not uniform across models or prompting strategies; CoT reduced but did not eliminate performance gaps. 
This suggests that multilingual reasoning challenges may stem from representational imbalance, tokenization inefficiencies, or weaker cross-lingual abstraction mechanisms rather than solely from insufficient training data volume. 
The results therefore contribute empirical evidence that translation-equivalent multilingual benchmarks may underestimate cognitive disparities in applied reasoning contexts.

Subject-specific variations further indicate that LLM competence is domain-sensitive rather than uniformly distributed. 
Higher performance in biology and descriptive chemistry contrasts with comparatively lower accuracy in physics and mathematically intensive domains. 
This pattern is consistent with previous findings that generative models perform strongly in knowledge retrieval and explanation tasks but encounter difficulty in multi-step symbolic or quantitative reasoning. 
However, the variability across prompting strategies—particularly the occasional underperformance of Few-Shot prompting relative to Zero-Shot—was unexpected. 
In several configurations, additional exemplars appeared to introduce noise or misalignment rather than clarity. 
This suggests that exemplar selection, formatting sensitivity, and contextual interference may meaningfully influence reasoning outcomes.
From a broader perspective, the findings underscore the limitations of general-purpose leaderboards in predicting domain-grounded academic competence. 
While models may achieve high scores on synthetic or short-form reasoning tasks, performance under authentic examination conditions reveals structural weaknesses in bilingual reasoning, subject transferability, and prompt sensitivity. 
The study therefore highlights the necessity of ecologically valid benchmarks that reflect operational educational realities.

Several limitations temper the conclusions. 
First, evaluation was restricted to multiple-choice questions, which constrain reasoning assessment to discrete answer selection and do not capture partial credit or procedural reasoning transparency. 
Second, reliance on API-based proprietary models introduces reproducibility dependencies on version stability and closed training data. 
Third, although bilingual, the framework currently includes only English and Hindi, limiting broader generalization across India’s linguistic diversity. 
Finally, exact-match scoring does not quantify reasoning quality independent of final answer correctness.
Future research should expand evaluation to subjective and long-form responses, incorporate diagrammatic and assertion-reason question formats, and extend coverage to additional Indian languages such as Tamil, Bengali, and Marathi. 
Further investigation into prompt optimization dynamics, cross-lingual tokenization effects, and calibration metrics would deepen understanding of bilingual reasoning gaps. 
Longitudinal benchmarking across model updates would also clarify whether observed disparities reflect transient architectural limitations or persistent structural constraints. 
Collectively, these directions would advance the development of more robust, linguistically equitable, and cognitively reliable large language models for educational deployment.

\section{Conclusion}
\label{sec:conclusion}
This study introduced IndicEval, a domain-grounded bilingual benchmarking framework designed to evaluate large language models using authentic Indian high-stakes examination questions. 
By moving beyond synthetic NLP tasks and incorporating real UPSC, JEE, and NEET assessments in English and Hindi, the framework captures subject complexity, multilingual reasoning demands, and procedural realism absent from conventional benchmarks. 
Empirical results show that Chain-of-Thought prompting yields substantial performance gains (mean improvement: $\mu = 11.59\%$, $\sigma = 9.47\%$), particularly in multi-step reasoning tasks. 
At the same time, significant multilingual degradation persists, with average cross-language accuracy gaps of 34.77 percentage points in lower-performing contexts. 
Subject-level variation and a negative correlation between descriptive and computational domain performance ($r = -0.67$, $p < 0.05$) further indicate uneven reasoning specialization across models.
Collectively, these findings demonstrate that high benchmark scores on generic leaderboards do not guarantee robust, bilingual academic competence. 
IndicEval contributes a reproducible, extensible infrastructure for ecologically valid evaluation and highlights the necessity of linguistically equitable and domain-sensitive benchmarking. 
Practically, the framework enables policymakers, educators, and developers to assess model readiness for deployment in multilingual educational environments. 
Theoretically, it reinforces the importance of context-aware evaluation for understanding reasoning reliability. 
The central takeaway is clear: advancing LLM capability requires not only larger models, but more rigorous, culturally grounded evaluation standards that reflect real-world cognitive demands.


\begin{thebibliography}{99}

\bibitem{wang2018glue}
Wang, A., Singh, A., Michael, J., Hill, F., Levy, O., Bowman, S.R.: GLUE: A multi-task benchmark and analysis platform for natural language understanding. In: \textit{arXiv preprint arXiv:1804.07461} (2018)

\bibitem{wang2019superglue}
Wang, A., Pruksachatkun, Y., Nangia, N., Singh, A., Michael, J., Hill, F., Levy, O., Bowman, S.R.: SuperGLUE: A stickier benchmark for general-purpose language understanding systems. In: \textit{Advances in Neural Information Processing Systems} (2019)

\bibitem{wang2021benchmarks}
Wang, Y., Zhang, H., Zhang, Y., Zhang, X.: Are NLP benchmarks misleading? A case study on GLUE. In: \textit{Proceedings of the 2021 Conference on Empirical Methods in Natural Language Processing} (2021)

\bibitem{zhang2023agievalhumancentricbenchmark}
Zhang, Z., Zhang, A., Li, M., Smyth, R., Banerjee, S., Zhang, Y., Cryan, J.: AGIEval: A human-centric benchmark for evaluating large language models. In: \textit{arXiv preprint arXiv:2304.06364} (2023)

\bibitem{liang2022holistic}
Liang, P., Bommasani, R., Lee, T., Tsipras, D., Sap, D., Chi, E., Hashimoto, T.: Holistic evaluation of language models. In: \textit{Transactions on Machine Learning Research} (2022)

\bibitem{zhang2023m3exameval}
Zhang, Y., Zhang, N., Ruan, L., Cao, K., Zheng, Z., Cai, Y., Hou, X., Lin, Y., Qian, Z., Zhou, J.: M3Exam: A multilingual, multimodal benchmark for evaluating large language models. In: \textit{Proceedings of the 61st Annual Meeting of the Association for Computational Linguistics} (2023)

\bibitem{clark2023sciexam}
Clark, P., Cowhey, S., Etzioni, O., Khot, T., Sabharwal, A., Schoenick, C., Tafjord, O.: Think you have solved question answering? Try ARC, the AI2 reasoning challenge. In: \textit{arXiv preprint arXiv:1803.05457} (2023)

\bibitem{kumar2023bharatibench}
Kumar, A., Bhatia, K., Talukdar, P., Joshi, P.: BharatiBench: A benchmark for Indian language NLP. In: \textit{Proceedings of the 13th Language Resources and Evaluation Conference} (2023)

\bibitem{singh2024indicqa}
Singh, R., Joshi, A., Kumari, M., Das, S.: IndicQA: A question answering dataset for Indian languages. In: \textit{Proceedings of the 2024 Conference of the North American Chapter of the Association for Computational Linguistics} (2024)

\bibitem{joshi2024culturalchasm}
Joshi, P., Singh, S., Kunchukuttan, A., Bhattacharyya, P.: The cultural chasm in LLM evaluation. In: \textit{Computational Linguistics} (2024)

\bibitem{patel2023languageswitching}
Patel, V., Sharma, A., Desai, S., Rao, P.: Language switching in competitive exam preparation: Implications for LLM evaluation. In: \textit{Proceedings of the 2023 EdTech India Conference} (2023)

\bibitem{mehta2024bilingualism}
Mehta, S., Rao, P.: The role of bilingualism in engineering entrance exams. In: \textit{Journal of Educational Technology}, vol. 28, no. 3, pp. 245--268 (2024)

\bibitem{nair2023beyondmultiplechoice}
Nair, A., Kumar, S., Saxena, P.: Beyond multiple-choice: Evaluating long-form reasoning in large language models. In: \textit{arXiv preprint arXiv:2310.08972} (2023)

\bibitem{srivastava2024exams}
Srivastava, A., Rastogi, A., Rao, B., Shoham, A.B., Abrego, G.H., Fiedel, G., Malkin, S., Nunez, A., Kassner, N., Kroiss, G., Levin, T., Machnev, M., Pan, J., Qmorris, D., Salimans, T., Sevcik, J., Simpson, A., Singhal, K., Sinha, T., Soliman, A., Song, X., Souza, G.F., Tanaka, S., Tsai, A., Wang, J., Wiegand, L., Yeo, J., Zhang, L., Zhuo, M., Zhuo, R., ZAreab, O.: EXAMS: A multi-subject benchmark for multilingual and code-mixed question answering. In: \textit{Proceedings of the 2024 Joint Conference on Lexical and Computational Semantics} (2024)

\end{thebibliography}
\end{document}